# Road Detection in Snowy Forest Environment using RGB Camera

*Sirawich Vachmanus, Takanori Emaru#, Ankit A. Ravankar#, Yukinori Kobayashi#

*Graduate School of Engineering, Hokkaido University, Sapporo, Japan

#Research Faculty of Engineering, Hokkaido University, Sapporo, Japan

Email: emaru@eng.hokudai.ac.jp, ankit@eng.hokudai.ac.jp

*Abstract* – Automated driving technology has gained a lot of momentum in the last few years. For the exploration field, navigation is the important key for autonomous operation. In difficult scenarios such as snowy environment, the road is covered with snow and road detection is impossible in this situation using only basic techniques. This paper introduces detection of snowy road in forest environment using RGB camera. The method combines noise filtering technique with morphological operation to classify the image component. By using the assumption that all road is covered by snow and the snow part is defined as road area. From the perspective image of road, the vanishing point of road is one of factor to scope the region of road. This vanishing point is found with fitting triangle technique. The performance of algorithm is evaluated by two error value: False Negative Rate and False Positive Rate. The error shows that the method has high efficiency for detect road with straight road but low performance for curved road. This road region will be applied with depth information from camera to detect for obstacle in the future work.

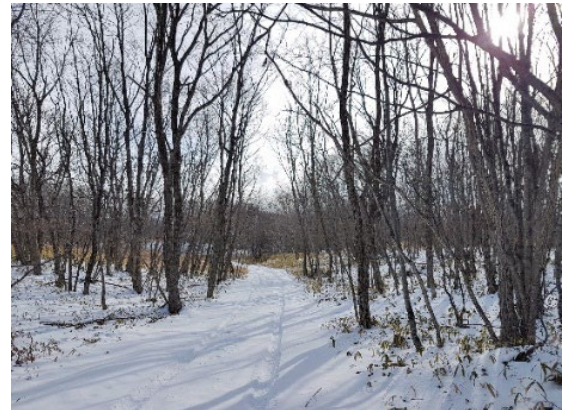

Fig 1. Environment of snowy forest road in Tomakomai city

## 1. Introduction

The autonomous driving system are widely used in many situations such as exploration, mapping, rescue, transportation etc. The system needs an effective sensor to detect the region of road and obstacles. Some of previous works [1] used laser scanner for road detection by elevating variance of information. The elevation variance represents the region of road and non-road by extracting plain surfaces. Other works [2] proposed the road detection and extraction by using sonic sensor. This method uses the range information of empty region to classify the road area.

Recently, the use of camera sensor for data collection from the environment has increased significantly due to their cost, and the availability of increased computation using embedded processors and computers. Some researcher [3] have developed filtering technique for use of vision in the dynamic weather environment for autonomous driving. The method of noise minimization for classification of road region is used in their proposed work.

In this paper, our target areas are snowy forest road as shown in Figure 1, which consist of snowy road, bushes, and trees. The experiment data set was collected in Tomakomai city, Hokkaido region in Japan. In this environment there are significant differences of intensity in object colour, so it is suitable for the visual method detection for road detection.

## 2. Literature Review

Previous researchers such as [3], proposed the 4 stages method for improving road detection system in outdoor environment with environmental noise. In the first stage, they extract the features of image by using various filters and select colour as main feature for road detection. Colour is represented in Red-Green-Blue (RGB) intensity from the camera and assume that road and non-road have the difference colour. But in noisy environment this feature gives low accuracy because various regions will be having comparable colour characteristics.

From poor extraction of road and non-road feature, the second stage is necessary. There are three filters used in this phase to detect and remove the environmental noise such as rain, snow, light intensity and shadow. In this section they convert the RGB colour space to HSV colour spaces because using of saturation (S) value and value (V) in this space it is easy to identify shadow. The shadow removing process is using of transformation function which is represented as $I_k^i(i,j)$ as shown in Eq. (1) where $\mu_{buff,k}$ and $\sigma_{buff,k}$ is the mean and variance of pixels of image $I$ at location $I_{buff,k}$, while $\mu_k$ is mean of shadow pixels of image $I$ at location $I_k$ [4].

$$I_k^i(i,j) = \mu_{buff,k} + \frac{I_k(i,j) - \mu_k}{\sigma_{buff,k}} \quad (1)$$

In the algorithm of rain and snow filtering, the method is based on modelling first the intensity value pixel of image. Eq. (2) describe intensity value of rain or snow image.

$$I_{rs} = \alpha I_E + (1-\alpha) I_b \quad (2)$$

$I_{rs}$ is the image intensity which is captured during rain or snow, $I_E$ is intensity of stationary drop of snowflakes, $I_b$ is background intensity, and $\alpha$ is ratio of time taken and time of exposure. By using Eq. (2) the rain and snow are removed but some information of edges with same colour with rain or snow are removed too. However, this lost information can be restored by using Eq. (3) to extract redefined guidance image ($I_g$).

$$I_g = \frac{I_f + J_g}{2} \quad (3)$$

where $J_g$ is gray image of $J_i - I_{rs}$ [5].

In the light filtering method, the image that is captured during sunlight, needs to minimize the light intensity. The reflection model is defined by Eq. (4).

$$I_{rs} = \alpha I_E + (1-\alpha) I_b \quad (4)$$

$$I(x) = I^D(x) + I^S(x) = w_d(x)B(x) + w_s(x)G$$

where, $I$ is image intensity, x = {x,y} is image coordinate, $I^D$ and $I^S$ is diffuse and reflection component, $B(x)$ is diffuse colour, $G$ is specular colour, $w_d(x)$ is coefficient of diffuse reflection component.

The method for minimizing the effects of intensity of light is expressed by Eq. (5).

$$I^D(\Lambda_{max}) = I - \frac{\max_{u \in (r,g,b)} I_u - \Lambda_{max} \sum_{u \in (r,g,b)} I_u}{1 - 3\Lambda_{max}} \quad (5)$$

where $I^D$ is minimal light intensity effect and $\Lambda_{max}$ is specular pixel.

In the segmentation stage use of Support Vector Machine (SVM) is popular because of improved performance in generalization ability [6]. The post processing stage is phase of morphological operation in the assumption of continuous road area.

In [7], vanishing point detection of the road is presented. The focus of their research is based on perspective images of urban streets. In the assumption of vanishing point is appear at intersection point of 2 parallel line in perspective image, they try to detect and track the vanishing point of that road.

The section of detection using line segments algorithm, three-line extraction methods are used in this part: [8], the probabilistic, and standard Hough transform. The method is found as the best in term of result number and geometric properties. The extracted lines are separated into 2 groups, horizontal line and vertical line for finding out the horizontal vanishing point. The best estimation of vanishing point is found by RANSAC [8]. For the intersection point of lines, the cross product of 2 lines is used. A line pair is regarded as an outlier if the angle between vanishing point from cross product and line pair is greater than pre-defined threshold. But this method is effective with road that lane mark is detected, however in snowy road all of lane make is covered by snow.

## 3. Methodology

Algorithm to develop road detection system, consist of 3 stages: filtering stage, morphological & classification stage, and vanishing point detection stage. Figure 2 show the flow of the algorithm.

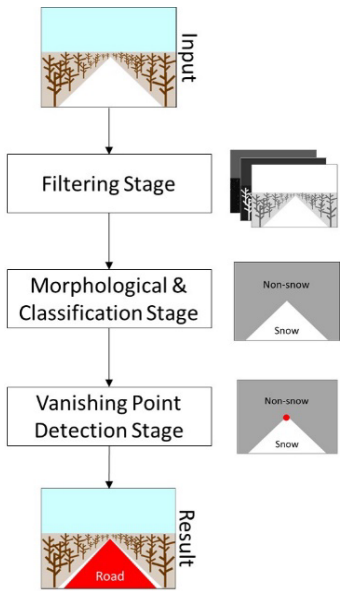

Fig 2. Road detection system

The input image is RGB image which is captured by omni stereo sensor which is installed on roof of car as show in Figure 3. An example of input image is shown in Figure 4.

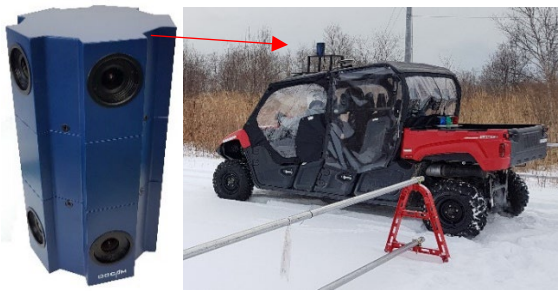

Fig 3. Omni Stereo and experimental car

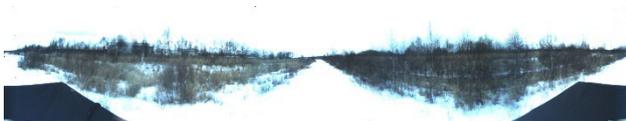

Fig 4a. RGB image from Omni Stereo

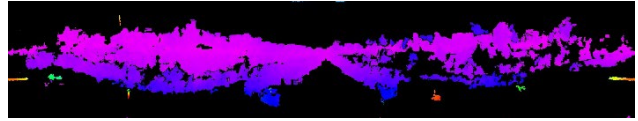

Fig 4b. Depth image from Omni Stereo

a) <u>Filtering Stage</u>

In the filtering stage noise is removed to enhance the image for extracting the snowy part. RGB colour space is converted to HSV colour space because of the easy identification of snow and tree bushes component. The method of converting colour space is presented by Eq. (6-8).

$$V = 255 \max(R, G, B) \quad (6)$$

$$S = \begin{cases} 255\frac{C}{V} & ; V \neq 0 \\ 0 & ; V = 0 \end{cases} \quad (7)$$

$$H = \begin{cases} \frac{30(G-B)}{C} & ; V = R \\ \frac{60 + 30(B-R)}{C} & ; V = G \\ \frac{120 + 30(R-G)}{C} & ; V = B \end{cases} \quad (8)$$

where $C$ is calculated by Eq. (9)

$$C = V - \min(R, G, B) \quad (9)$$

$R, G, B$ are component of red, green, blue in RGB colour space. $H, S, V$ are component of Hue, Saturation, Value in HSV colour space.

In HSV colour space, the other filter is applied such as Gaussian filter, Histogram equalization, and Light filter [3] to enhance the image for classification.

b) <u>Morphological & Classification Stage</u>

In this stage each component of environment is extracted such as snow and non-snow part. Because of high contrast of each part in the image, the classical image processing method is applied to classify the component. The morphological transformations are used in this section to remove the effect of noise like a small snow spot in bushes.

c) <u>Vanishing point detection</u>

To detect the road covered by snow and to find out the actual road is impossible by camera. The small part of snow is scoped as a region of interest (ROI) and assumed as road area. This ROI is defined by the vanishing point and size of car. The detection of vanishing point is done using fitting triangle method as show in Figure 5, in the assumption of road in perspective image is show as triangular shape. In Figure 5, green area is the detected region of snow from classification stage and red line is the fitting triangle of green region. The top position of fitting triangle is defined as the vanishing point.

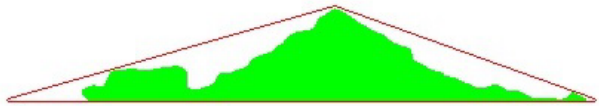

Fig 5. Fitting triangle

## 4. Experimental Evaluation and Result

In this paper, the detection of road in snowy forest environment by using of Omni Stereo camera was tested. By utilizing methods as described in Section 3, the experimental results are shown in Figure 8.

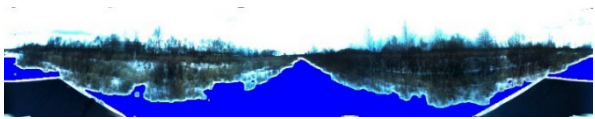

Fig 6. Snowy area

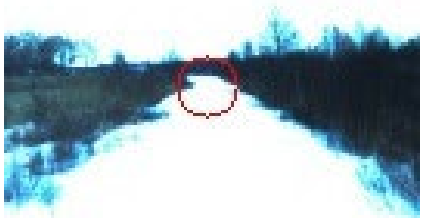

Fig 7. Vanishing point

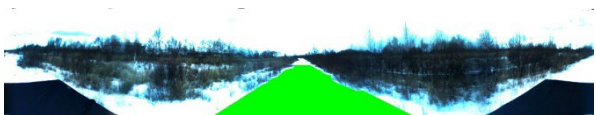

Fig 8a. Road area

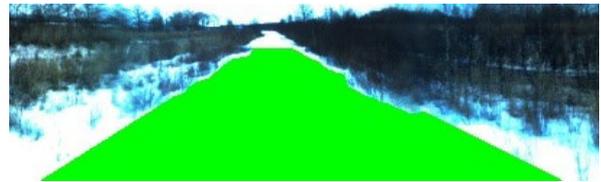

Fig 8b. Zoom image of (6a)

The snow detection result is presented by blue region as shown in Figure 6. By using fitting triangle method, the vanishing point is found as centre of red circle in Figure 7 and the road area is show by green region in Figure 8a and 8b.

The system evaluation of the experiment is based on ground truth classification. Eq. (10-11) are used for the performance factor of system.

$$FNR = \frac{FN}{TP + FN} \qquad (10)$$

$$FPR = \frac{FP}{TN + FP} \qquad (11)$$

where FNR represent false negative rate, FPR represent false positive rate, FN is number of road pixel which are detected as non-road, FP is number of non-road pixel which are detected as road, TP is number of road pixel which are detected as road, TN is number of non-road pixel which are detected as non-road.

The using this evaluation technique to determine the error of the proposed system, 20 sample images from same environment were tested. The error rate of the test are shown in Figure 9. FNR shows the miss-detection of road area and FPR show the faulty-detection of non-road area. The average FNR of 20 sample is about 6.2% and for FPR is about 3.2%.

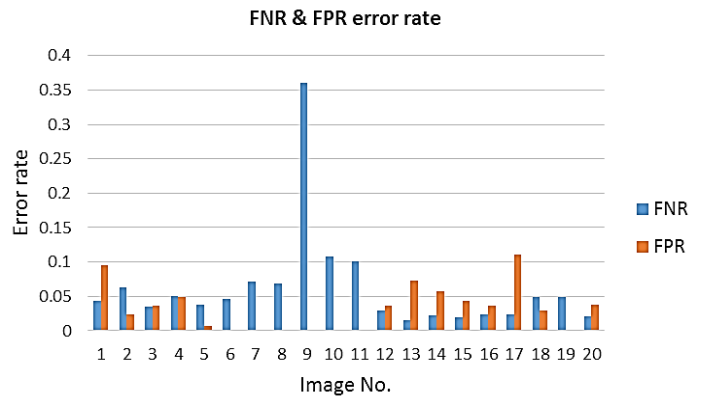

Fig 9. FNR & FPR error rate

In the case of curved road, this algorithm is not supported. Because of this method is using the assumption of perspective direct road is triangle, the curve shape is not identified. In the future work the using of path centre determination [9] will be applied to develop this method for detecting the curved road.

## 5. Conclusion

This study introduces the method of road detection system. The environment of detecting road is in a snowy forest environment. The technique consists of filtering step which enhance the component of image and remove the effect of noise, the morphology stage is then utilized for better result of classification. The vanishing point detection system is used of fitting triangle technique for scope into the vanishing point of image. However, the snowy road detection system can find the part of road which covered by snow, but for the road with junction or more than one vanishing point the system will generates low efficiency. In the future this detected road area will adapt with depth information fin figure 4b for detect the obstacle on the road and applied some method for curve road detection. Furthermore, other methods such as perspective transform can be utilized to get the bird's eye view of the scene to recognize lane boundaries and curvature of the road. We plan to implement these methods as future work.

**Acknowledgment**

This work was supported by "Project to support the upgrading of strategic core technology", The Hokkaido Bureau of Economy, Trade, and Industry, Japan. And supported by Tomakomai-City and Tomatoh Inc. in carrying out the experiments.